\documentclass{article}

\usepackage{arxiv}

\usepackage[utf8]{inputenc} 
\usepackage[T1]{fontenc}    
\usepackage{hyperref}       
\usepackage{url}            
\usepackage{booktabs}       
\usepackage{amsfonts}       
\usepackage{nicefrac}       
\usepackage{microtype}      
\usepackage{lipsum}		
\usepackage{graphicx}
\usepackage{natbib}
\usepackage{doi}
\usepackage{nicefrac}       
\usepackage{microtype}      
\usepackage{xcolor}         

\usepackage{algorithm}
\usepackage{algpseudocode}
\usepackage{amsmath}
\usepackage{mathtools} 
\usepackage{booktabs} 
\usepackage{tikz} 
	
\usepackage{color, colortbl}
\usepackage{tabularx}
\usepackage{algpseudocode}
\usepackage{multirow}

\usepackage{graphicx}
\usepackage{wrapfig}
\usepackage{lipsum}
\usepackage{footnote}
\usepackage{caption}   
\usepackage{graphicx}  
\usepackage{caption}
\usepackage{subfigure} 
\usepackage{multicol}
\usepackage{wrapfig}
\usepackage{float}

\title{Actively learning a Bayesian matrix fusion model with deep side information}

\author{
  Yangyang Yu \& Jordan W. Suchow\\
  School of Business\\
  Stevens Institute of Technology\\
  Hoboken, NJ 07030 \\
  \texttt{\{yyu44,jws\}@stevens.edu} \\
}

\renewcommand{\shorttitle}{\textit{arXiv} Template}

\hypersetup{
pdftitle={A template for the arxiv style},
pdfsubject={q-bio.NC, q-bio.QM},
pdfauthor={David S.~Hippocampus, Elias D.~Striatum},
pdfkeywords={First keyword, Second keyword, More},
}

\begin{document}
\maketitle

\begin{abstract}
High-dimensional deep neural network representations
of images and concepts can be aligned to
predict human annotations of diverse stimuli. However,
such alignment requires the costly collection
of behavioral responses, such that, in practice, the
deep-feature spaces are only ever sparsely sampled.
Here, we propose an active learning approach to
adaptively sampling experimental stimuli to efficiently
learn a Bayesian matrix factorization model
with deep side information. We observe a significant
efficiency gain over a passive baseline. Furthermore,
with a sequential batched sampling strategy,
the algorithm is applicable not only to small
datasets collected from traditional laboratory experiments
but also to settings where large-scale
crowdsourced data collection is needed to accurately
align the high-dimensional deep feature representations
derived from pre-trained networks.
\end{abstract}


\section{Introduction}

In cognitive research, Bayesian probabilistic models typically serve two principal roles: one as a hypothesis positing how individuals draw inferences from their observations of the environment, and the other as a tool enabling scientists to learn from observations of human behavior (\cite{vul2014one, griffiths2008bayesian, mamassian2002bayesian}). Our work acts as an intermediate approach that bridges these two uses of Bayesian models. We use Bayesian Probabilistic Matrix Factorization (BPMF) with deep side information to align a machine representation of entities to human behavioral responses to those entities, such that the model serves as both a model of people's mental representations and as a predictive model of their behavior. This method enables accurate inference of behavioral responses by generating low-rank predictions of perceptual response matrices. It offers a viable model structure to align machine vision systems with human visual perception. For instance, it can integrate the bimodal information from facial imagery and psychological attributes and yield predictions of people's impressions of human faces. 

BPMF has proven effective in consolidating multi-source information and predicting missing responses while constructing confidence intervals (\cite{salakhutdinov2008bayesian, adams2010incorporating}). In traditional laboratory experiments, it has been successfully applied to predict individuals' perceptual outcomes based on human interpretable features (\cite{zhang2020active}). 

When implementing BPMF in large-scale behavioral prediction tasks that involve object features from multiple modalities, two significant challenges arise. The first challenge is that high-quality prediction demands advanced machine-generated features that outperform traditional human-defined ones. Machine learning algorithms have the capacity to generate a vast number of new objects based on previously perceived ones, enhancing diversity and realism while reducing bias among stimuli. However, using deep learning algorithms such as generative adversarial networks (GANs) (\cite{kammoun2022generative}) to extract high-dimensional, informative features from various objects increases computational costs nearly linearly with the number of data features. This poses a challenge to the integration of BPMF with deep learning methods, making it practical only to sparsely sample from the deep-feature space. Second, there is a lack of crucial information needed for making reliable predictions due to the highly sparse response matrix. This issue becomes especially problematic in human-subjects research, where data collection operates under constrained budgets and resources, deals with a large participant population, and may have an extensive list of questions. Budgetary constraints may limit the number of questions asked, and lengthy experimental instruments could lead to inaccurate responses due to participants resorting to mental shortcuts (\cite{krosnick1991response}). Hence, large-scale experiments can often only collect responses from a small fraction of the total instrument from each participant. To effectively implement BPMF with deep side information on large-scale behavioral prediction tasks, it is therefore crucial to develop a data sampling strategy that can effectively target the most informative data points, enabling the achievement of satisfactory predictive results despite these constraints.

Active learning is a data acquisition technique that can interactively identify the most informative samples to efficiently create a training data set. Although this training set can be compact, it possesses a powerful predictive capacity. Active learning has been widely employed to tackle problems associated with accuracy in sparse matrix completion (\cite{elahi2016survey, 6729492}). One of its key strengths is the capacity to accurately infer the complete response distribution from a limited selection of samples, obviating the need to query the majority of the response matrix.

Here, we introduce a method for active learning in a BPMF model, utilizing uncertainty sampling (\cite{sugiyama2006active}) and k-Center Greedy selection (\cite{sener2017active}). We demonstrate that this method more efficiently learns a multimodal model compared to a passive baseline. We further investigate the impact of varying MCMC approximation simulation chain lengths on the active sampling performance to optimally integrate active learning into the BPMF model framework. Estimating the posterior uncertainty brings a cost associated with the number of posterior MCMC samples collected. A trade-off exists between a slow yet precise estimate and a quick but less accurate one. We investigate this trade-off by manipulating the number of MCMC samples used for the posterior uncertainty estimation in model parameters and measuring its effect on the overall algorithm performance under a fixed computational budget.

In this paper, we first explain the framework for actively learning the deep Bayesian matrix factorization model. We then train the model on a large behavioral dataset --- the One Million Impressions dataset (\cite{peterson2022deep}) --- and validate the Bayesian model learning efficiency as well as performance improvement by applying a suitable active strategy to acquire a more informative training pool.
Finally, we demonstrate that our method achieves promising predictive results by adaptively querying a minority of the data for model training.

\section{Related work}
\label{sec2}

We discuss related work in three separate sections below, on the topics of using Bayesian probabilistic matrix factorization (BPMF) for human behavioral prediction, high-dimensional deep feature representations as the side information of BPMF and active learning.

\subsection{Bayesian Probabilistic Matrix Factorization}

BPMF acts as a bridge between human cognition and mathematical inductive inference. Based on assumptions regarding cognitive problems, it integrates prior knowledge from multiple channels, guiding participants to update their beliefs using mathematical forms. Moreover, it dictates how participants update their beliefs in response to new data with mathematical forms. In the context of visual perception modeling, BPMF uses a matrix structure to combine 2-way side information - visual objects and observers' features, over priors. Through Monte Carlo Markov Chain (MCMC) simulations, Bayesian posteriors are computed and linked to perceptual inference outcomes (\cite{mamassian2002bayesian, kersten2004object}). Two components in the MCMC simulation chain are warm-up steps and posterior samples. Warm-ups initialize model parameters so that the Markov chain can efficiently sample from the neighborhood of substantial posterior probability mass and travel in equilibrium. Posterior samples, formed after warm-up, shape the posterior distribution and generate predicted outcomes. A balance between the number of warm-ups and posterior samples is needed to avoid significant fluctuations and overkill in MCMC simulation performance (\cite{mcelreath2020statistical}). It helps the BPMF achieve satisfactory predictive results through MCMC.

\subsection{Modeling Perception Using Deep Features}

Most human perceptual inference tasks involve combining information from multiple modalities, such as visual and linguistic information. Previous research (\cite{peterson2022deep, zhang2018unreasonable}) has demonstrated the superiority of machine-generated deep features over conventional human-interpretable attributes (e.g., color or size in images). This is primarily due to 1) the comprehensive nature of high-dimensional deep object representations, and 2) the ability to self-generate new objects, which is crucial for predicting novel objects in the model system. Pretrained networks from deep learning domains, such as StyleGAN2 (\cite{karras2020analyzing}) for images and Sentence-BERT (\cite{reimers2019sentence}) for text, can generate high-quality object feature sets for cognitive studies, thus reducing considerable data-collection costs. Hence, within the BPMF paradigm, deep features serve as a more effective choice than human-interpretable features for generating predictions.

\subsection{Active Learning}
Taking inspiration from sparse matrix completion research (\cite{settles2009active, chakraborty2013active}), we leverage active learning as a practical approach to impute a full response distribution for each participant over diverse objects, given a response matrix with limited entries. This approach prioritizes data points for querying in the response matrix, allowing for a faster alignment of machine inferences with human thinking with fewer queried data points. Active learning accelerates this alignment by adaptively choosing the most informative data samples, a critical approach in large-scale cognitive experiments with constrained resources. In our model, two active strategies are deployed to enhance BPMF performance with a limited training set: 1) Uncertainty sampling (\cite{chakraborty2013active,sutherland2013active}), prioritizing data points with the highest predictive uncertainties in each iteration as the most informative; 2) k-Center Greedy sampling (\cite{sener2017active}), seeking k data points that maximize the mutual information between them and the remaining unselected data pool in every iteration. This is achieved by pre-setting k centers and identifying data points that minimize the maximum distance of these points to the nearest centers.

\section{Methods}
\label{sec:methods}
Our approach improves the performance of perceptual learning on limited training cases by implementing the active strategy to the Bayesian matrix factorization with deep side features. The details of this approach and notation are articulated in two sections below.

\subsection{Deep Bayesian Probabilistic Matrix Factorization}
\label{sec:featuresBPMF}
First, in predicting first impressions, a pre-trained deep network is used to create face and trait features. Utilizing indices $j$ for face images, $h$ for traits, and $i$ for participants, Peterson et al. (\cite{peterson2022deep}) proposed a method for extracting deep face features $f_j$ using StyleGAN2 generator. Meanwhile, we employ the Sentence-BERT model to create a deep trait feature set $t_h$. This becomes in two latent spaces for the response matrix $R$: a 512-dimensional image feature latent space $\mathit{\mathcal{F}}$ for faces and a 300-dimensional linguistic feature space $\mathit{\mathcal{T}}$ for traits. These spaces are represented by conditioned, unit-variance, multivariate normal latent variables: $\omega_{f_{j}}$ for the face space and $\omega_{t_{h}}$ for the trait space. A parallel experiment involved lower-dimensional dense deep features, conducted by further processing pre-trained image and trait features through a multi-layer perceptron (MLP) model (\cite{yu2022deep}). The goal is to determine whether encapsulating deep features to a lower dimensionality provides a sufficient representation and if it's necessary to include an intermediate dimension reduction operation before inputting side features into the Bayesian model. If so, understanding the appropriate degree of dimension reduction is key to ensuring adequate feature representativeness while minimizing computational cost and simulation time.

Second, under the BPMF setting, the latent variables represent the computational coefficients for the participants' impression ratings $R_{jh}$. The coefficients are estimates based on two priors of $\omega_{f_{j}}, \omega_{t_{h}} $ following spherical Gaussian distributions (\cite{gurkan2022cultural}). In Formulas~\ref{eqn:eq1} and \ref{eqn:eq2}, the two-sided features are fused together after being converted into the same dimensionality in the latent space, and then the responses $\hat{R}_{jh}$ are predicted through simulation.

\begin{equation}
  \begin{split}
p(\omega_{f_{j}}|\sigma) = \text{Normal}(\omega_{f_{j}} | \sigma^{-1} \mathbf{I}); \quad p(\omega_{t_{h}}|\sigma) = \text{Normal}(\omega_{t_{h}} | \theta^{-1} \mathbf{I}),
 \end{split}
 \label{eqn:eq1}
  \end{equation}
where $\sigma$ and $\theta$ are independent and gamma distributed.
\begin{equation}
  \begin{split}
&F_{j} =  f_{j} \times \omega_{f_{j}}^{T} ;\quad \quad \quad \; \; \quad T_{h} =  t_{h} \times \omega_{t_{h}}^{T}; \\
&R^{*}_{j h} = F_{j} \times T_{h}^{T} + \varepsilon_{j h}; \quad \hat{R}_{j h} = \text{sigmoid}(R^{*}_{j h}) \times 100 ,
 \end{split}
 \label{eqn:eq2}
  \end{equation}
where $f_{j}$ and $t_{h}$ are row vectors and $R^{*}_{j h} \in (-\infty, \infty)$ is projected into predicted ratings as a continuous value $\hat{R}_{j h} \in (0, 100)$.

Third, we further analyze the impact of the simulation process on our deep BPMF model's performance. By running parallel simulation chains of equal length but varying proportions of warm-ups to posterior samples, we identify the optimal mean for impression inference. Furthermore, we employ sequential Markov chain Monte Carlo (MCMC) (\cite{yang2013sequential}) as the approximation method for modeling BPMF across the whole dataset. This choice is guided by the MCMC method family's suitability for large hierarchical model computation (\cite{banerjee2003hierarchical}), and its effectiveness in our case for integrating over thousands of parameters associated with deep features. We observe that aggregating parameters from the last ten Bayesian posteriors rather than all posteriors stabilizes model predictions, reducing sampling randomness. Compared to traditional MCMC, sequential MCMC more readily extracts the last few posteriors in each iteration, offering reduced memory usage and faster computation when using Numpyro in Python as the development platform.



\subsection{Actively Learning The Matrix Factorization Model}
\label{sec:activeBPMF}

By implementing active learning strategies in the training data sampling process of BPMF, we aim to use as few filled matrix entries as possible in training, predict all the rest in the response matrix, and converge the Bayesian model in fewer iterations. We uniformly choose a very small initial training pool $\boldsymbol{S^{0}}$ with size $L$ (5, 8, etc.), selected randomly from the whole rated data pool $\boldsymbol{S}$, which has size $N$. Notice that only the initial pool is accessible to us. The feature pair $(f_{jn}, t_{hn})$ of an arbitrary data entry $\mathcal{C}_n$ in $\boldsymbol{S}$ is shortened as $\mathcal{U}_n ,\,  n \in \{1, ..N\}$, and its rating as $R_{in}$. The feature pair of a data entry $\mathcal{C}_l$ in the initial pool $\boldsymbol{S^{0}}$ is denoted as $\mathcal{U}_l$ with observable rating as $R_{il}$, $l \in \{1,...,L \}$. Besides the initial pool, we assume $Q$ opportunities as the budget for asking an oracle for information about an extra $p$ data points and a learning algorithm $A_{\boldsymbol{S}}$ to guide us in choosing the appropriate points at each time. $A_{\boldsymbol{S}}$ generates an updated set of parameters $\{\omega_{f_{j}}^{*},\omega_{t_{h}}^{*}\}$ based on $\boldsymbol{{S^{0}}^{*}} = \boldsymbol{S^{0}} \cup \boldsymbol{S^{p}}$. The choice of $\boldsymbol{S^{p}}$ for the oracle to label is a subset of $\boldsymbol{S}$, which minimizes the future expected learning loss. Formula~\ref{eqn:eq4} shows the future expected loss after $q$ times label querying.

\begin{equation}
  \begin{split}
&\underset{\boldsymbol{S^{p}_{q}}}{\text{min}} \, E_{(\mathcal{C}_1,...,\mathcal{C}_N) \sim \boldsymbol{S}}[\mathcal{L}(\mathcal{C}_{n}; A_{\boldsymbol{S^{0}} \cup ... \boldsymbol{S^{p}_{q}}})]
 \end{split}
 \label{eqn:eq4}
  \end{equation}
Compared to the classic definition of active learning, our active strategy has two significant differences: 1) We set $A_{\boldsymbol{S}}$ to query a batch of $p$ data points in each round instead of a single data point because previous works (\cite{sener2017active, zhang2020active}) illustrate that with a large dataset and high-dimensional deep features, the performance improvement made by one data point in each round is negligible. 2) As active learning has mainly been shown to be effective for tasks with discrete prediction targets (\cite{settles2009active, yona2022active}), we further expand its capability of handling continuous prediction cases by reformulating the learning metrics and loss functions. Here, we explore two types of batched active strategies to construct and minimize the loss function for continuous targets and compare their performance. One is pure uncertainty-based sampling, and the other takes the trade-off of sampling diversity and uncertainty of sampling into account. 

\subsubsection{Uncertainty Sampling}

One classic approach of uncertainty selection is targeting samples with the least confidence in predictions in each active iteration. Sugiyama and Ridgeway (\cite{sugiyama2006active}) justified selecting data points with the maximum standard deviations of predictive distribution, $\sigma_{{\omega_{\mathcal{U}_{l}}^{*}}}$ (shorten as $\sigma$),  as the most uncertain samples in the context of Gaussian linear regression. Here, we further extend this strategy to Bayesian matrix factorization because of two critical similarities: 1) like their work, we have a continuous prediction target, and 2) the prediction distribution is also a univariate Gaussian distribution,  $\rho(R^{*})$.  

In this paper, we validate the performance of uncertainty-based active Bayesian matrix factorization on the impression prediction task. Algorithm \ref{alg:uncertainty} and Formula \ref{eqn:eq5} (in Appendices \ref{uncertain_formula}) present how this strategy works iteratively to select the $p$ most uncertain samples as a batch to add to the training pool.

\begin{algorithm}[htb]
  \caption{Batched Uncertainty Sample Selection for Deep Bayesian Matrix Factorization}
  \label{alg:uncertainty}
  \begin{multicols}{2}
  \begin{algorithmic} 
    \Require
      The set of randomly selected $d$ ratings as initial pool $\boldsymbol{S^{0}}$;
      The rating $R_{il}$ in $\boldsymbol{S^{0}}$ is given by a participant $i$ for a face image in terms of a certain trait. And budget $Q > 0$. 
    \Ensure
      Predicted ratings for all entries of the response matrix, $\hat{R}$
    \State Extract deep features face images $f_{jl}$ and trait $t_{hl}$
    \label{code:fram:extract pretrained}
    \Repeat \quad $Q$ times, initialize $q = 1$; 
    \State Update parameters ${\omega_{f_{j}}^{*}}_{q},{\omega_{t_{h}}^{*}}_{q}$ via MCMC 
    \label{code:fram:mcmc}
    \State Compute predicted $\hat{R}_{q}$, $\sigma_{q}$ for the response matrix
    \label{code:fram:metric}
    \State Query $\boldsymbol{S^{p}_{q}}$  with $\text{argmax} \sum_{z=1}^{p}  \; {\sigma_{z}}_{q} \footnotemark$ 
    \label{code:fram:query}
    \State $\boldsymbol{S^{0}_{q}} \gets \boldsymbol{S^{p}_{q}} \cup \boldsymbol{S^{0}_{q-1}} $ 
    \label{code:fram:union} 
    \State q = q + 1
    \label{code:updateIterator}
    \Until{$q = Q$}\\
   \Return $\hat{R}_{Q}$, $\boldsymbol{S} \setminus \boldsymbol{S^{0}_{Q}}$
  \end{algorithmic}
 \end{multicols}
\end{algorithm}

\subsubsection{k-Center Greedy Sampling}

k-Center Greedy sampling is a pool-based active sampling strategy. Sener et al. (\cite{sener2017active}) regard active learning loss for classifications using CNN given a batch of $p$ samples, $E_{(\mathcal{C}_1,...,\mathcal{C}_N) \sim \boldsymbol{S}}(\mathcal{L}(\mathcal{C}_{n}; A_{\boldsymbol{S^p}}))$, as containing three components: generalization loss, training loss, and core-set loss. $p$ denotes the number of central points in the unexplored data pool and also equates to the count of points chosen during each adaptive sampling cycle.

In our scenario, which involves predicting continuous targets, the first two losses are primarily managed by the Bayesian model, while the active learning strategy concentrates on minimizing core-set loss. Core-set loss is defined as the distance between average empirical loss over the points with known ratings and that over the entire dataset, including entries with unknown ratings. Sener et. al shows its upper bound as $\mathcal{O}(d_{\boldsymbol{S^0} \cup \boldsymbol{S^p}}) + \mathcal{O}(\sqrt{\frac{1}{N} })$. Consequently, the optimization goal of loss can be converted to minimize the coverage radius from $p$ centers ($d_{{\boldsymbol{S^0} \cup \boldsymbol{S^p}}}$), as illustrated in  Figure~\ref{fig:kcenter_fig_v1} (in Appendices~\ref{k-center_graph} and \ref{kcenter_details}). The minimization of $d_{{\boldsymbol{S^0} \cup \boldsymbol{S^p}}}$ is further deduced as  $\underset{\boldsymbol{S^p}}{\text{min}}\, \underset{\mathcal{C}_n \in \boldsymbol{S} \setminus (\boldsymbol{S^0} \cup \boldsymbol{S^p})}{\text{max}}\,\underset{\mathcal{C}_l \in \boldsymbol{S^0} \cup \boldsymbol{S^p}}{\text{min}} \, \bigtriangleup (\mathcal{U}_n, \mathcal{U}_l) $ when $p$ data points are queried.This formula can be computed based on the distances of feature pairs $\mathcal{U}_n = (f_{jn}, t_{hn})$ and $\mathcal{U}_l = (f_{jl}, t_{hl})$ in a two-dimensional coordinate system.
The $\underset{\boldsymbol{S^p}} {\text{min}}\,{\boldsymbol{S^0} \cup \boldsymbol{S^p}}$ is iteratively recomputed in Algorithm \ref{alg:greedy}  to search for new data points to be queried.

\begin{algorithm}[htb]
  \caption{k-Center Greedy Sample Selection for Deep Bayesian Matrix Factorization}
  \label{alg:greedy}
  \begin{multicols}{2}
  \begin{algorithmic} 
    \Require
      The set of randomly selected $d$ ratings as initial pool $\boldsymbol{S^{0}}$;
      the rating $R_{il}$ in $\boldsymbol{S^{0}}$ is given by a participant $i$ for a face image in terms of a certain trait. And budget $Q > 0 $. 
    \Ensure
      Predicted ratings for all the entries of the response matrix, $\hat{R}$
    \State Extract deep features face images $f_{jl}$ and trait $t_{hl}$
    \label{code:fram:extract pretrained2}
    \Repeat \quad $Q$ times, initialize $q = 1$; 
    \State Update parameters ${\omega_{f_{j}}^{*}}_{q},{\omega_{t_{h}}^{*}}_{q}$ via MCMC 
    \label{code:fram:mcmc2}
    \State  Set $p$ learning centers 
    \label{code:fram:centers}
    \State Choose $p$ centers $({\mathcal{C}_1}_q, ..., {\mathcal{C}_p}_q) = S^{p}_q$ \footnotemark by   \\
  \quad $\boldsymbol{S^{p}_q} =  \underset{{\mathcal{C}_n}_z \in  S\setminus {S_{z}^{p}}_{q}}{\text{argmax}} \;  \text{min}_{{\mathcal{C}_{l}}_z \in {S_{z}^{p}}_{q}} \;\bigtriangleup({\mathcal{U}_{n}}_z, {\mathcal{U}_{l}}_z )$,\\
      where set $\boldsymbol{S^{p}_q} \subseteq \boldsymbol{S}  \setminus \boldsymbol{S^{0}_{q-1}}$\\
    \label{code:fram:optimzation}
    \State $\boldsymbol{S^{0}_{q}} \gets \boldsymbol{S^{0}_{q-1}} \cup \boldsymbol{S^{p}_q}$
    \label{code:fram:union1} 
    \State q = q + 1
    \label{code:updateIterator2}
   \Until{$q = Q$}\\
   \Return $\hat{R}_{Q}$, $\boldsymbol{S} \setminus \boldsymbol{S^{0}_{Q}}$
  \end{algorithmic}
\end{multicols}
\end{algorithm}

\footnotetext[1]{${\sigma_{z}}_{q}$ denotes the predicted standard deviation for the $z$th queried data point in a batch $p$ samples during the $q$th iteration of the uncertainty strategy}
\footnotetext[2]{${\boldsymbol{S}_{h}^{p}}_{q}$ denotes the $k$th queried data point in a batch $p$ in the $q$th iteration of k-Center Greedy active strategy.}

\section{Dataset}


The One Million Impressions dataset (\cite{peterson2022deep}) is a large-scale dataset containing people's perceptual judgments based on their first impressions of human faces. This dataset contains over 1 million ratings of 34 traits for 1,000 machine-generated face images. Each face image is rated by 30 unique individuals for each trait. Given that each image only receives ratings from a limited group of participants, the response matrix is relatively sparse. This dataset is ideal for our approach as it allows for the integration of deep image and trait attributes as side information in BPMF.

\section{Experiments and Results}
\label{sec:experiments}

We conduct three experiment sets to explore the applicability of implementing the active learning strategy in the training data sampling process for deep Bayesian matrix factorization model. This approach comprises active learning, deep input features, and BPMF. Each experiment tweaks one component. The active learning method's efficacy and efficiency are examined by the Root Mean Square Error (RMSE). The test RMSE, computed on the dataset excluding the queried training samples, is the most critical metric, reflecting the model's predictive ability for unknown data. The models' evaluation metrics are averaged over three experimental repetitions.

\begin{figure*}[htbp]
  \centering
  \includegraphics[width=14cm, height=4.9cm]{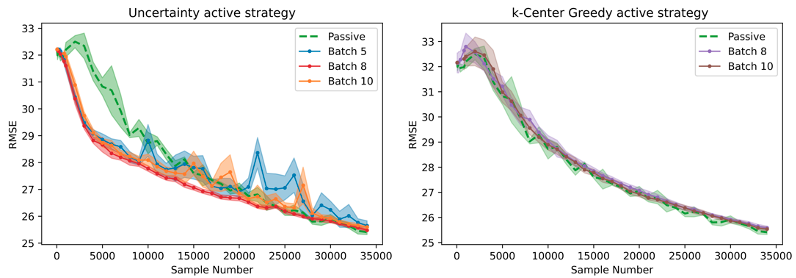}
  \caption[linedot]{Test RMSEs over sample number with different batch sizes for uncertainty vs. k-Center Greedy active strategies with $95\%$ confidence intervals.\footnotemark}
  \label{linedot}
\end{figure*}

\subsection{Choices of Strategy and Batch Size}

The first experiment seeks to identify the optimal batch size for an active learning task on a large scale. We employ two active sampling strategies, uncertainty and k-Center Greedy, with deep Bayesian matrix factorization models to predict first-impression ratings. These strategies adaptively query batch samples, with batch size varying per experimental setting. Five active models with distinct strategy and batch size combinations are run concurrently. The initial training pool, equivalent in size to the chosen batch size, is randomly drawn from the unknown data pool. The baseline is the same deep Bayesian matrix factorization model with passive sampling that randomly selects increasing numbers of samples across iterations, without adaptive querying. Pretrained StyleGAN2 image features and Sentence-BERT trait features serve as the deep BPMF's input side information.

\begin{wrapfigure}{r}{7cm}
  \centering
  \includegraphics[width=6.8cm, height=5.0cm]{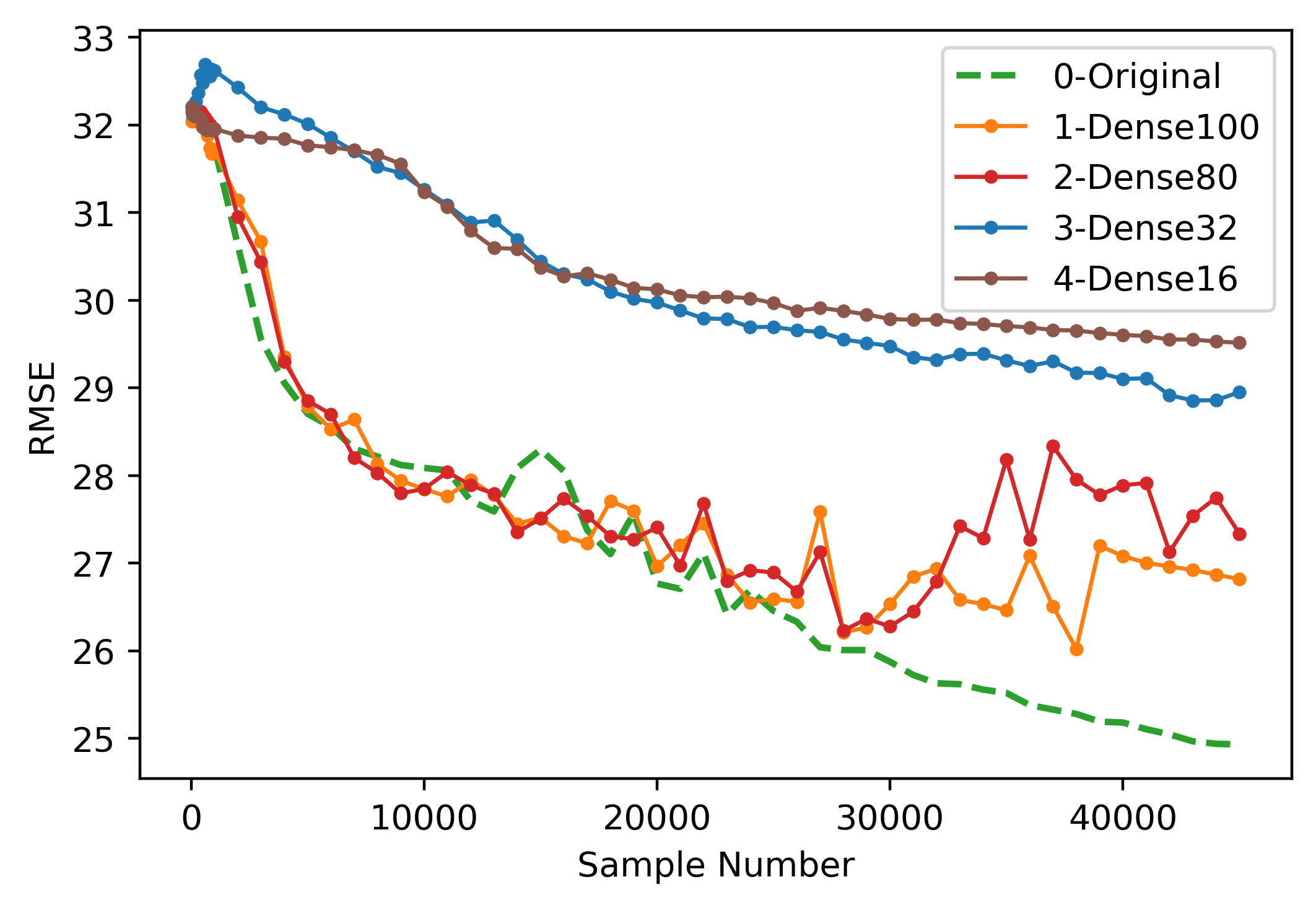}
  \caption{Test RMSEs over sample number for the  BPMF with batch-10 uncertainty active strategy under different choices of input features dimensions.}
  \label{dense_summary}
\end{wrapfigure}

Figure \ref{linedot} presents adaptive sampling across 34,000 samples, where the uncertainty active strategy surpasses both the k-Center Greedy strategy and the passive baseline model. Among the uncertainty sampling batch sizes, a batch size of 8 emerges as the optimal choice, demonstrating the fastest and sharpest decrease in test RMSE, and experiencing minimal fluctuations and the narrowest confidence interval as the training pool size increases. Its advantage is particularly notable when the training pool has fewer than 10,000 samples, representing roughly $0.1\%$ of the entire dataset. In this case, the test RMSE drops sharply from 32.1 to 27.8, approximately 1.0 lower than the passive learning's test RMSE. The uncertainty strategy with a batch size of 10 achieves the second most efficient convergence of the six models but exhibits less stability in test RMSE during the intermediate training pool size in a range of [15,000, 25,000]. The batch size 5 model encounters severe fluctuations when the training pool grows larger than 20,000 and sometimes has noticeably higher predicted RMSEs and a wider confidence interval than the baseline model, indicating it is the worst model in the uncertainty strategy group.  On the contrary, regardless of training set size, neither batch-8 nor batch-10 k-Center Greedy performs better than the baseline model. After the training pool reaches 28,000 samples, models with batched uncertainty active sampling gradually converge with the baseline model at a test RMSE of about 25.8, indicating the training pool has reached a size of containing extensive information. Thus, even random selection works as effectively as the active strategy due to the Bayesian model's estimation and control for future predictive uncertainty. Consequently, we confirm that the batched uncertainty active strategy for deep Bayesian matrix factorization can efficiently decrease predictive RMSE when training data size is limited. Furthermore, the choice of batch size influences the model's prediction stability.

\footnotetext[3]{To smooth out fluctuations in RMSEs, each dot in Figures \ref{linedot} and  \ref{dense_summary} is averaged from 200 samples of the adjacent epochs.}

\subsection{Efficacies of Input Feature Dimensionalities }

The second set of experiments probes the influence of different deep-feature dimensionalities on deep Bayesian matrix factorization models with uncertainty active sampling. We investigate the trade-off between feature dimensionality and predictive performance and determine the optimal set of feature vectors. We use the original pre-trained deep features (512-dimensions for images, 300 for traits) as inputs for the benchmark model (Model 0). Models 1--4 incorporate dense feature layers, from the bimodal MLP model (\cite{yu2022deep}), with reduced dimensionalities of 100, 80, 32, and 16 before fusion.

Figure~\ref{dense_summary} shows that while Model 0 consistently retains a leading position in terms of test RMSE, Models 1 and 2, using 100- and 80-dimensional dense features respectively, achieve comparable test RMSEs when the training pool size is small to moderate (up to 30,000 samples). However, when the sample size reaches 30,000, both Models 1 and 2 exhibit significant fluctuations in RMSE, suggesting their inability to capture sufficient informative samples for future iterations. Models 3 and 4, using 32- and 16-dimensional features, fail to target informative data points conductive for learning performance irrespective of training pool size. The models' performance indicates that retaining approximately $33\%$ of trait features and $20\%$ of face features' dimensionality enables the active strategy to sample informative points when the training set is sparse. However, the ability to query effectively degrades when the training set is more enriched. We infer that the loss of subtle feature information due to the nonlinear feature transformation of dimension reduction is the primary reason for this. The consistently high test RMSEs for Models 3 and 4, with the lowest input dimensions, support this inference. As the feature dimensionality decreases, the information loss worsens from negligible to critical.

\begin{table}[H]
\centering
\begin{tabular}{p{3.0cm} p{2.8cm} p{3.3cm} p{2.8cm}}
MCMC chain length \raggedright & Learning Type \raggedright& Running time (minutes) & Test RMSE \\
 \hline
220 & Active & 32.4633 & 27.024   \\
280 & Active  & 36.1294 & 26.1506   \\  
64,000 & Passive   & 29.3237 & 27.4982\\
80,000 & Passive   & 35.8137 & 27.7274\\
88,000 & Passive   & 39.5863 & 27.1459\\
\end{tabular}
\caption{Active learning vs. passive learning training time and test RMSE in the third experiment.}
\label{traingtime}
\end{table}

\subsection{Impact of Simulation Chain Length}

\begin{figure}[htbp]
\centering  
\subfigure[Test RMSEs for different model options over the lengths of the simulation chain with 350 training samples]{   
\begin{minipage}{7.05cm}
\centering    
\includegraphics[scale=0.186]{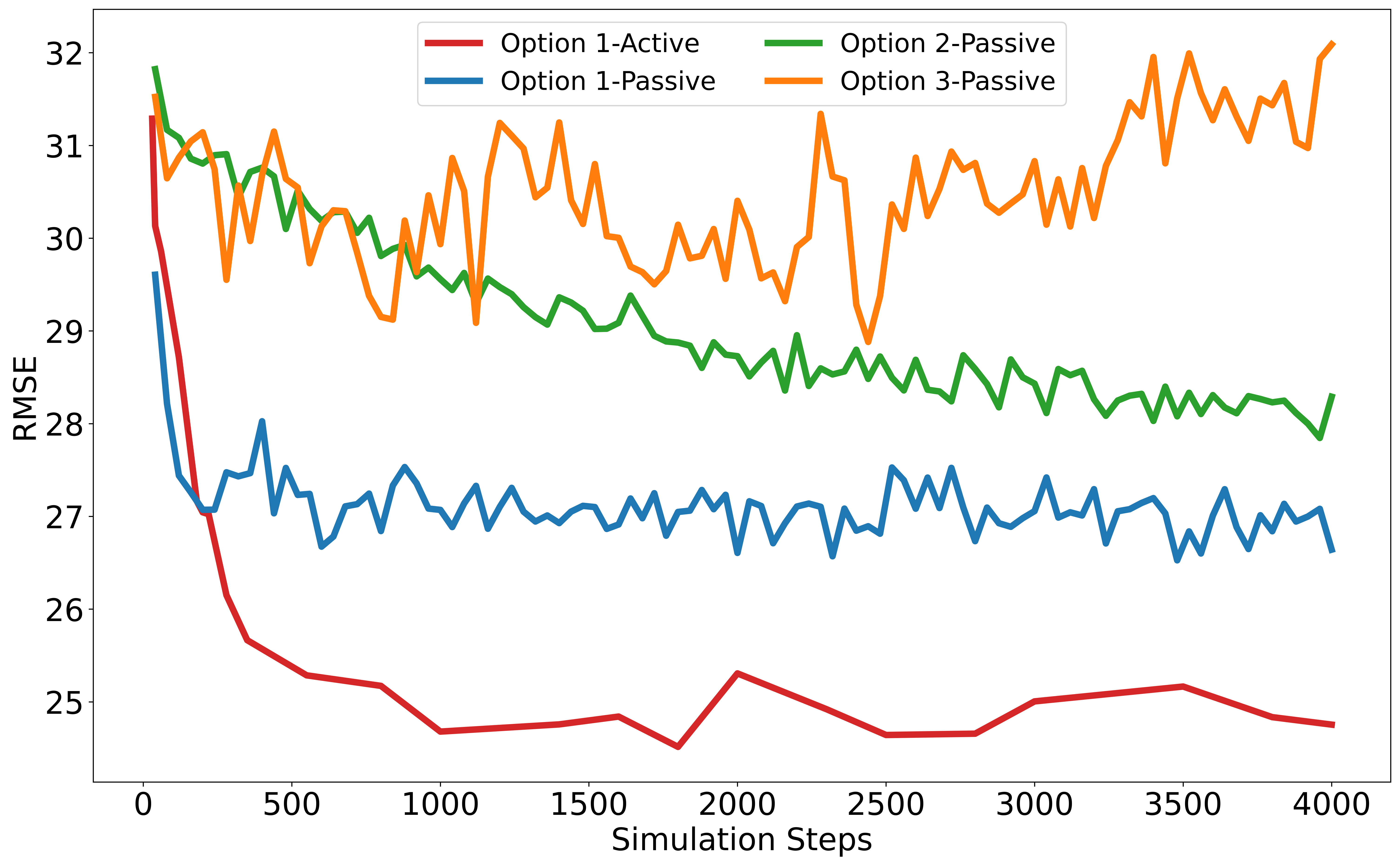}  
\end{minipage}
}
\subfigure[Test RMSEs for the BPMF using batch-2 active strategy with different simulation chain lengths over training sample numbers.]{ 
\begin{minipage}{7.05cm}
\centering    
\includegraphics[scale=0.186]{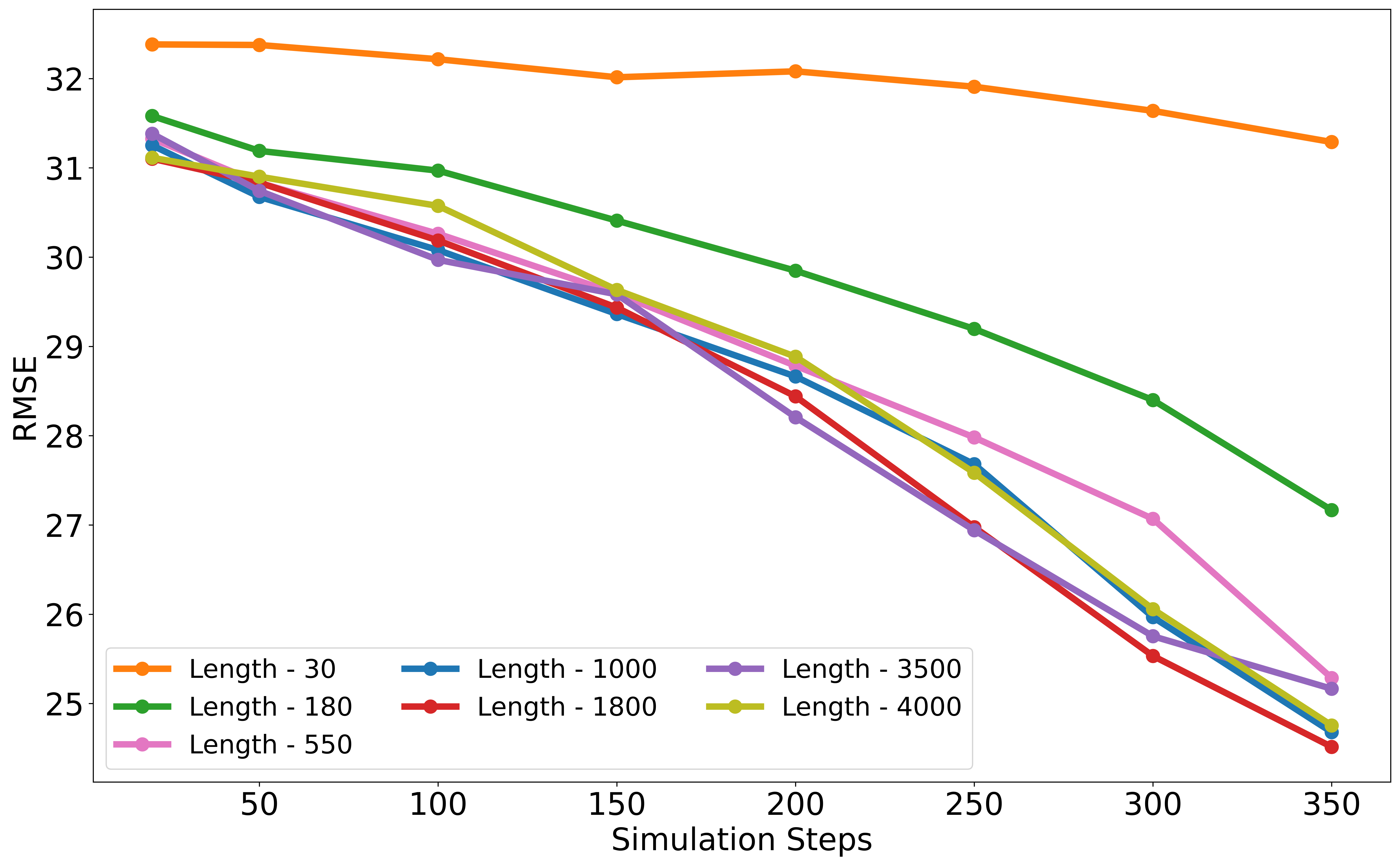}
\end{minipage}
}
\caption[exp3summary]{Result summary of the third experiment. \footnotemark}    
\label{exp3_fig}  
\end{figure}

\footnotetext[4]{To smooth out fluctuations in RMSE values, the dots in Figures~\ref{exp3_fig} are averaged from 5 adjacent epochs.}

In the third set of experiments, we explored three configurations for the Markov Chain Monte Carlo (MCMC) simulation chains: (1) increasing both warm-ups and posterior samples in a 3:5 ratio, in increments of 8 samples, (2) expanding the posterior samples by steps of 5 with no warm-ups, and (3) raising the warm-ups in steps of 5 with a constant 5-step posterior samples. After identifying the optimal configuration, we applied a batch-2 uncertainty active sampling strategy to the BPMF model. Through the comparison between active versus passive sampling on BPMF, we aimed to evaluate whether increasing the MCMC simulation chain's length can compensate for the efficiency advantage gained by active learning. For this, we compared the prediction efficiency of the deep BPMF with active versus passive sampling as the simulation chain lengthened. Given computational limitations, these experiments utilized a 1000-sample subset of the One Million Impressions dataset. This subset consisted of ten arbitrary responses from each combination of ten faces and ten traits, selected randomly beforehand. To control the outcome unpredictability caused by insufficient training data and to distinguish performance differences across experimental settings, we set the training pool to 350 samples and assessed model performance by comparing the predictive Root Mean Squared Error (RMSE) on the remaining 650 test data points. Given the considerable reduction in the total size of the dataset, the batch size for the active strategy was decreased to two.

We first executed three passive-sampling BPMF models, each with different simulation chain options, and compared their test RMSE trends over time (Figure~\ref{exp3_fig}(a)). Our observations revealed that the MCMC chains from Option 1 offered the highest reduction efficiency in predictive RMSE as the simulation steps increased. Options 2 and 3 consistently yielded higher test RMSEs than Option 1, demonstrating that a simulation chain with proportional increases in warm-ups and posteriors yields the most powerful predictions for human impression ratings. Moreover, we applied the optimal simulation chain (Option 1) to the BPMF model with a batched uncertainty active sampling strategy. Its results in Figure~\ref{exp3_fig}(a) showed a significantly better performance with fewer simulation steps than all passive learning models, except during the initial stages ($<$200 steps). By the time 220 simulation steps were reached, the active model's predictive RMSE dropped to 27.024, already 0.05 lower than the best-performing passive model with much longer simulation chains. When the simulation chain length extended to 1,800 steps, the active model achieved the best test RMSE of 24.515, 2.533 lower than the passive model with the same number of steps. After that point, the active model's test RMSE increased slightly but remained the lowest among all settings. As too few simulation steps lead to incorrect predictions in practice, we demonstrate that deep Bayesian matrix factorization with active strategy retains a sustained advantage over passive sampling BPMF on model performance. The training time analysis Table~\ref{traingtime} further illustrates that the advantage of active learning cannot be offset by extensively extending the simulation chain in terms of both training time and predictive performance. With a reasonably short simulation chain length of 220,  the active model's RMSE decreases to a level that the passive models can't reach. This training time analysis was run on a single NVIDIA GeForce RTX 4090 GPU with 24 GB memory.

Upon querying 350 samples, the Bayesian model with active strategy achieves optimal performance with a 1,800-step simulation chain. We investigate whether this simulation chain length can maintain the performance advantage throughout the adaptive sampling process. Figure~\ref{exp3_fig}(b) displays test RMSEs across varying training pool sizes for different simulation chain lengths. The test RMSE for the 1,800-step model remains among the lowest as the training set expands from 50 to 350 samples, confirming that the optimal performance achieved with 350 training samples is not an exceptional case for the model with 1,800-step simulation during the active sampling process.

\section{Conclusion and Discussion}
\label{sec:conlcusion}

In this paper, we propose and evaluate a method for actively learning a model that fuses deep neural network representations using Bayesian probabilistic matrix factorization to produce a predictive model of human first impressions. The active strategy identifies the most informative stimulus-attribute pairs to sample, considering the uncertainty in the posterior of the model parameters. Empirical experiments demonstrate its superior performance compared to passive learning. This advantage is crucial in real-world crowdsourced cognitive studies with limited experimental budgets. Additionally, our methodology exhibits broad applicability across numerous domains, such as in the development of online recommendations for social media, e-commerce, or behavioral prediction systems. With minimal data collected from a small segment of existing customers or users, our approach efficiently generates high-quality behavioral forecasts for new users or objects, thereby offering a solution to the prevalent cold-start problem in recommendation systems.

Four aspects related to the three main components of deep Bayesian matrix factorization using an active learning strategy jointly impact performance and computational efficiency on large behavioral datasets: the sampling strategy and batch size in active learning, the dimensionality of the deep feature space, and the lengths of the simulation chain in BPMF. Our first experiment with human impression data demonstrated that the uncertainty sampling strategy significantly outperforms the k-Center Greedy strategy. The uncertainty strategy selects samples based on predicted standard error computed from the posterior distribution over the latent feature space, while the k-Center strategy targets samples covering the diversity of the original pretrained deep features. With a small training pool, Bayesian inference is still capable of learning from the least confident data retrieved by the previous sampling iteration and extracts valuable information to efficiently reduce model predictive error in the next iteration, benefiting the uncertainty strategy. In contrast, the k-Center Greedy sampling struggles to choose cluster centers from input features, which reflect the centers of similar groups in the unknown pool. The choices of 8 or 10 centers fail to capture feature diversity, causing the k-Center Greedy strategy to underperform against passive learning. Choosing an appropriate batch size can enhance learning efficiency for active sampling. However, this decision is dependent on the size of the unknown data pool and the strategy mechanism. Although the unknown pool size is usually beyond our control, the first experiment showed that an active strategy accessing the Bayesian latent space, instead of initial input features, makes it easier to determine a feasible batch size.

Our second experiment tested whether reducing the dimensionality of deep features can optimize computational resources. The findings suggested that while reducing feature dimensionality can economize computational resources, it may impair model performance. This emphasized the importance of high-dimensional pretrained deep features over low-dimensional, human-interpretable features. We observed that reducing deep feature dimensionality by over $60\%$ was tolerable for a small training set. Consequently, the required degree of dimension reduction and prospective training set size need careful investigation when optimizing computational resources.

The impact of the MCMC simulation chain on the deep Bayesian matrix factorization model with active strategy is investigated from multiple perspectives in the third set of experiments. The outcomes indicate that the batched active sampling strategy empowers the Bayesian model with a shorter simulation chain, outperforming the passive models in predicting human first impressions. This underscores our assertion that an appropriate active strategy and batch size can enhance the performance of the deep BPMF, particularly when constrained by data query budgets and computational resources.

Despite the promising performance of deep Bayesian matrix factorization combined with the uncertainty active strategy in predicting people's impressions, two aspects of our model require further attention: 1) Our experiment utilized deep image and trait features as bimodal side information to predict impression ratings through Bayesian factorization, generating a 2D response matrix. Considering that Bayesian factorization can also handle 3D tensors (\cite{xiong2010temporal}), we are interested in exploring the generalizability of our approach to datasets with more complex structures and diverse features. For instance, in predicting people's impressions, a third information type, such as participants' features, could be included alongside image and trait attributes.
2) We expanded two classic active strategies initially designed for classifying discrete targets to predict continuous rating values. Our study confirmed the effectiveness of the uncertainty strategy and examined potential issues in the k-Center Greedy strategy. To further enhance learning efficiency,  the integration of the Bayesian model with more advanced adaptive sampling strategies is worth thorough investigation.

\section{Appendices}
\appendix    
\section{Queried data samples in k-Center greedy active learning strategy}
\label{k-center_graph}
\begin{figure}[htbp]
  \centering
  \includegraphics[width=0.5\linewidth, height = 4.5cm]{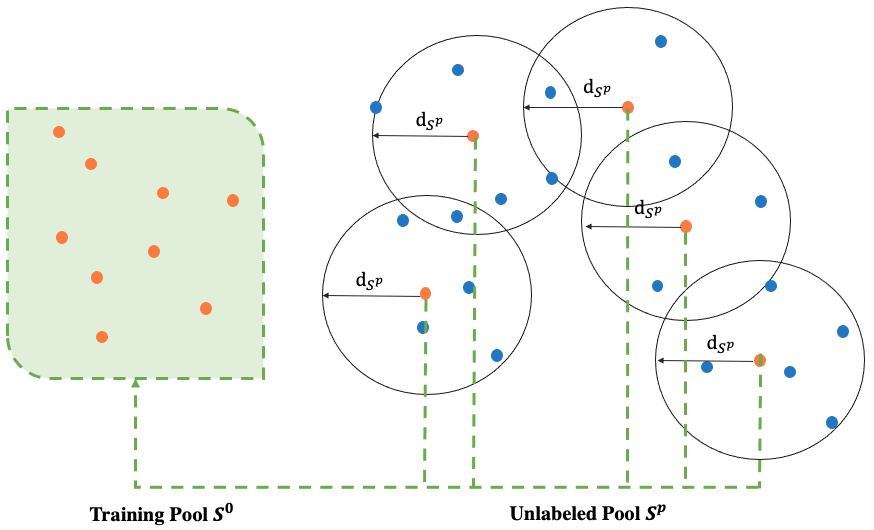}
  \caption{Queried data points and $d_{{S^0 \cup S^p }}$ in one batch.}\label{fig:kcenter_fig_v1}
\end{figure}

The following graph Figure~\ref{fig:kcenter_fig_v1} depicts $p$ centers and their associated coverage radii $d_{{\boldsymbol{S^0} \cup \boldsymbol{S^p}}}$ during one batch selection of k-Center Greedy active sampling. This representation is utilized to illustrate one essential component of the upper bound of Core-set loss.

\section{Batched sample selection formula of uncertainty active learning strategy}
\label{uncertain_formula}
\begin{equation}
  \begin{split}
&\boldsymbol{{S^{p}}^{*}}= \underset{\mathcal{C}_n \in  \boldsymbol{S} \setminus \boldsymbol{S^{p}} }{\text{argmax}} \, \sum_{h=1}^{p}  \; \sigma_{{\omega_{\mathcal{U}_{l}}^{*}}_{q}} (R^{*}|\rho(R^{*}))
 \end{split}
 \label{eqn:eq5}
  \end{equation}

\section{Extension of the mathematical details to formulate K-center Greedy active learning strategy}
\label{kcenter_details}
The subsequent Formulas~\ref{eqn:eq6}, ~\ref{eqn:eq7}, and ~\ref{eqn:eq10} illustrate further deductions of the batch selection criteria, informed by the knowledge that the upper bound of the Core-set loss primarily consists of the component $\mathcal{O}(d_{\boldsymbol{S}^0 \cup \boldsymbol{S}^p})$.
The task of finding the upper bound for Formula~\ref{eqn:eq6} is equivalent to the k-Center problem presented in Formula ~\ref{eqn:eq7}, which is described as the min-max facility location problem in Wolf's work (\cite{wolf2011facility}).
\begin{equation}
  \begin{split}
&\frac{1}{N}  {\textstyle \sum_{n \in \boldsymbol{S}} \mathcal{L} (\mathcal{C}_n; A_{\boldsymbol{S}^0 \cup \boldsymbol{S}^p} )}  - \frac{1}{L} {\textstyle \sum_{l \in \boldsymbol{S}^p} \mathcal{L} (\mathcal{C}_l; A_{\boldsymbol{S}^0 \cup \boldsymbol{S}^p} )} \\
&\le \mathcal{O}(d_{\boldsymbol{S}^0 \cup \boldsymbol{S}^p}) + \mathcal{O}(\sqrt{\frac{1}{N} } ) 
 \end{split}
 \label{eqn:eq6}
  \end{equation}

\begin{equation}
  \begin{split}
&\underset{\boldsymbol{S}^p}{\text{min}}\, d_{\boldsymbol{S}^0 \cup \boldsymbol{S}^p} \simeq   \underset{\boldsymbol{S}^p}{\text{min}}\, \underset{\mathcal{C}_n \in S\setminus (\boldsymbol{S}^0 \cup \boldsymbol{S}^p) }{\text{max}}\,\underset{\mathcal{C}_l \in \boldsymbol{S}^0 \cup \boldsymbol{S}^p}{\text{min}} \, \bigtriangleup (\mathcal{U}_n, \mathcal{U}_l)
 \end{split}
 \label{eqn:eq7}
  \end{equation}

\begin{equation}
  \begin{split}
&\boldsymbol{{S^{p}}^{*}} = \sum_{l=1}^{p} \; \underset{\mathcal{C}_l \in  \boldsymbol{S}\setminus \boldsymbol{S}^{p} } {\text{argmax}}\;  \text{min}_{\mathcal{C}_{L} \in \boldsymbol{S}^{p}} \;\bigtriangleup(\mathcal{U}_{l}, \mathcal{U}_{L}) 
 \end{split}
 \label{eqn:eq10}
  \end{equation}

\clearpage
\bibliographystyle{unsrtnat}
\bibliography{references}  






\end{document}